%% file: main_CameraReady.tex
\documentclass[10pt,twocolumn,letterpaper]{article}

\usepackage{cvpr}              % To produce the CAMERA-READY version
\input{preamble}
\definecolor{cvprblue}{rgb}{0.21,0.49,0.74}
\usepackage[pagebackref,breaklinks,colorlinks,allcolors=cvprblue]{hyperref}
\usepackage{pifont}
\usepackage{amssymb}
\usepackage{bbding}
\usepackage{multirow}
\usepackage{makecell}
\usepackage{float}
\usepackage[ruled,vlined]{algorithm2e}

\def\paperID{} % *** Enter the Paper ID here
\def\confName{CVPR}
\def\confYear{2026}

\title{
Towards Motion Turing Test: Evaluating Human-Likeness in Humanoid Robots
\vspace{-4mm}
}

\author{Mingzhe Li$^{1,2\ast}$, Mengyin Liu$^{1,2\ast}$, Zekai Wu$^{1,2}$, Xincheng Lin$^{1,2}$, Junsheng Zhang$^{1,2}$, Ming Yan$^{1,2,3}$, \\ Zengye Xie$^{1,2}$, Changwang Zhang$^{4}$, Chenglu Wen$^{1,2}$, Lan Xu$^5$, Siqi Shen$^{1,2\dagger}$, Cheng Wang$^{1,2}$ 
\\
$^1$Fujian Key Laboratory of Urban Intelligent Sensing and Computing, Xiamen University\\
$^2$Key Laboratory of Multimedia Trusted Perception and Efficient Computing,\\ Ministry of Education of China, School of Informatics, Xiamen University\\
$^3$National Institute for Data Science in Health and Medicine, Xiamen University\\
$^4$OPPO Research Institute, $^5$ShanghaiTech University\\
{}
\vspace{-22mm}
\and
\\
{}
}
 
\begin{document}
\maketitle

    \begingroup
    \renewcommand\thefootnote{\fnsymbol{footnote}}
    \footnotetext[1]{Equal contribution.}
    \footnotetext[2]{Corresponding author.}
    \endgroup

%%%%%%%%% ABSTRACT

 \input{sec/0_abstract}

%%%%%%%%% BODY TEXT

 \input{sec/introduction}

 \input{sec/related}

 \input{sec/dataset}

\input{sec/benchmark}
 \input{sec/experiment}

\vspace{-2mm}
\section{Conclusion}
\label{sec:conclusion}
In this work, we introduce the Motion Turing Test and HHMotion dataset, a comprehensive benchmark of humanoid and human motions annotated by 30 annotators. Despite advances, robots still lag in complex actions. Our PTR-Net baseline effectively predicts motion human-likeness, outperforming existing VLM models and serving as a tool for evaluation and reinforcement learning. Together, they provide a rigorous, human-centered foundation for developing more natural and expressive humanoid motions.
\paragraph{Acknowledgment.} This work was partially supported by the Fundamental Research Funds for the Central Universities (No. 20720230033), by Xiaomi Young Talents Program. We would like to thank the anonymous reviewers for their valuable suggestions.

% \printbibliography
%-------------------------------------------------------------------------
%%%%%%%%% SUPPLEMENTARY TEXT

% \input{sec/supplementary}

{
    \small
    \bibliographystyle{ieeenat_fullname}
    \bibliography{main}
}

% WARNING: do not forget to delete the supplementary pages from your submission 
% \input{sec/X_suppl}

\end{document}

%% file: sec/0_abstract.tex
\begin{abstract}
Humanoid robots have achieved significant progress in motion generation and control, exhibiting movements that appear increasingly natural and human-like. Inspired by the Turing Test, we propose the \textbf{Motion Turing Test}, a framework that evaluates whether human observers can discriminate between humanoid robot and human poses using only kinematic information. To facilitate this evaluation, we present the \textbf{H}uman-\textbf{H}umanoid \textbf{Motion} (\textbf{HHMotion}) dataset, which consists of 1,000 motion sequences spanning 15 action categories, performed by 11 humanoid models and 10 human subjects. All motion sequences are converted into SMPL-X representations to eliminate the influence of visual appearance. We recruited 30 annotators to rate the human-likeness of each pose on a 0–5 scale, resulting in over 500 hours of annotation. Analysis of the collected data reveals that humanoid motions still exhibit noticeable deviations from human movements, particularly in dynamic actions such as jumping, boxing, and running. Building on HHMotion, we formulate a \textbf{human-likeness evaluation task} that aims to automatically predict human-likeness scores from motion data. Despite recent progress in multimodal large language models, we find that they remain inadequate for assessing motion human-likeness. To address this, we propose a simple \textbf{baseline} model and demonstrate that it outperforms several contemporary LLM-based methods. The dataset, code, and benchmark will be publicly released to support future research in the community.
\end{abstract}
\vspace{-4mm}

%% file: sec/introduction.tex
\section{Introduction}
\label{sec:intro}
\begin{figure}[htb]
    \centering
    \includegraphics[width=0.8\linewidth]{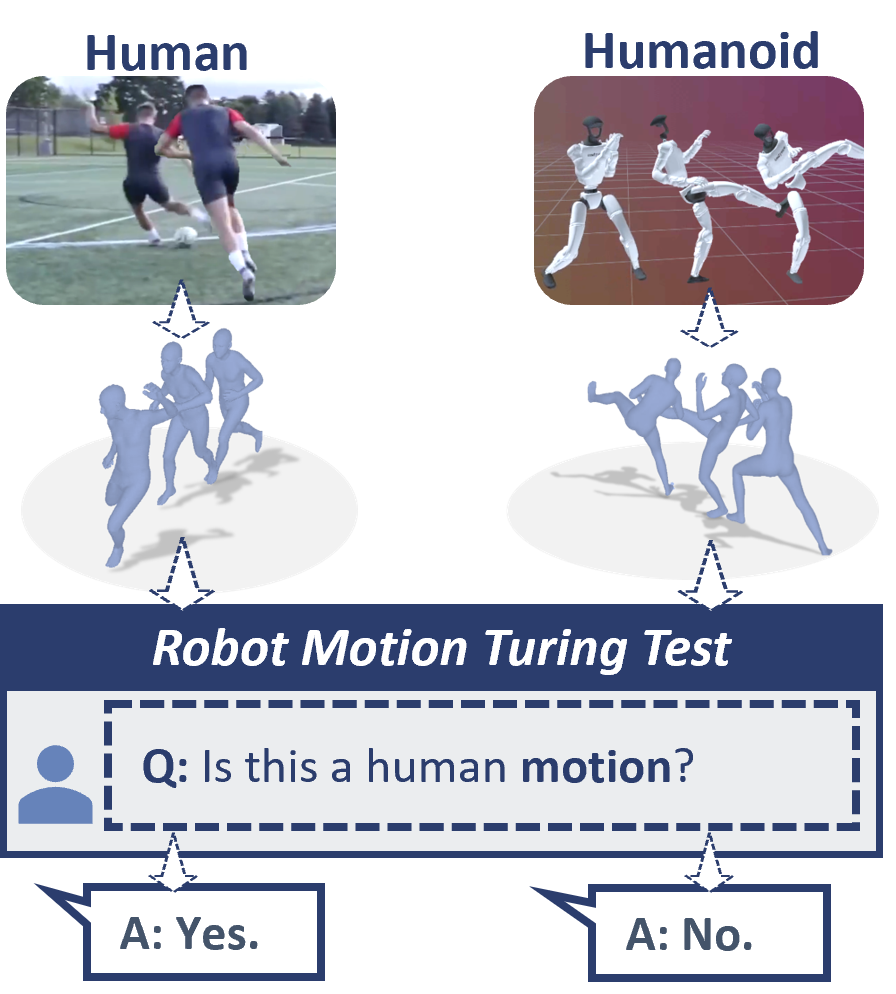}
    \caption{\textbf{Motion Turing Test}: Evaluators judge whether the pose sequence resembles human motion, focusing solely on motion without appearance cues.}
    \label{fig:intro}
    \vspace{-5mm}
\end{figure}
Research on humanoid robot motion has made remarkable progress in motion generation~\cite{allshire2025videomimic, fu2024humanplus, tang2024humanmimic, okami2024, harmon2024, yang2025omniretarget, he2024learning, he2024omnih2o, mao2024learning, baek2024human} and motion control~\cite{he2025hover, fan2025one, chen2024learning, gu2024humanoid, CassieJumpLi2023, radosavovic2024real}, enabling increasingly natural, stable, and human-like movements. At recent robotics conferences and events~\cite{whrg2025, waic2025, wrc2025}, humanoid robots have been observed performing actions such as walking, running, dancing, and even gymnastics motions with impressive fluidity. 

\begin{table*}[t]
     % \vspace{-4mm}
     \caption{\textbf{Comparisons with related datasets.} The ``-" symbol indicates that it is not included in the dataset. The ``*" symbol denotes that this specific data type in the dataset is a combination of prior datasets and does not incorporate newly acquired data.}
     \vspace{-2mm}
     \centering
     \resizebox{1\linewidth}{!}{
     \begin{tabular}{lcccccccccccc}
        \toprule[1pt]

        \multirow{2}{*}{Dataset} & \multicolumn{2}{c}{Human} & \multicolumn{2}{c}{Humanoid} & 
        \multirow{2}{*}{\makecell{\#Humanoid\\Models}} &
        \multirow{2}{*}{Human\&Humanoid} & 
        \multirow{2}{*}{Human-likeness score} & 
        \multirow{2}{*}{Duration} & 
        \multirow{2}{*}{Motion} & 
        \multirow{2}{*}{\#Category} & 
        \multirow{2}{*}{\#Clips} \\

        \cline{2-5}
        
          & MoCap & Video & Simulation & Real     \\

        \midrule
        AMASS~\cite{AMASS_ICCV2019} & \CheckmarkBold* & - & - & - & - & - & - & 40+h & General & - & 26.3k \\
        Human3.6M~\cite{Ionescu14a} & \CheckmarkBold & \CheckmarkBold & - & - & - & - & - & - & Daily & 15 & - \\
        LAFAN1~\cite{harvey2020robust} & \CheckmarkBold & - & - & - & - & - & - & 4.6h & Sports & - & - \\
        Motion-X~\cite{lin2023motion} & \CheckmarkBold  & \CheckmarkBold & - & - & - & - & - & 144.2h & General & - & 81,084 \\
        IDEA400 (Motion-X)~\cite{lin2023motion} & \CheckmarkBold & \CheckmarkBold & - & - & - & - & - & - & General & 400 & 13k \\
        SMART~\cite{chen2021sportscap} & \CheckmarkBold & \CheckmarkBold & - & - & - & - & - & - & Sports & - & 5000 \\
        DHRP~\cite{heo2024diverse} & - & - & \CheckmarkBold & \CheckmarkBold & 11 & - & - & - & Daily+Sports & - & 462 \\
        HumanoidRobotPose~\cite{amini2021real} & - & - & - & \CheckmarkBold & 10+ & - & - & - & Football & - & 23 \\
        PHUMA~\cite{lee2025phuma} & \CheckmarkBold* & \CheckmarkBold & \CheckmarkBold & - & 2 & \CheckmarkBold & - & 72.96h & General & 11 & 76.01k \\
        LAFAN1 Retarget~\cite{lafan1_retarget_dataset} & - & - & \CheckmarkBold & - & 3 & - & - & - & Sport & 8 & 120 \\
        AMASS Retarget~\cite{Retargeted_AMASS_R} & - & - & \CheckmarkBold & - & 1 & - & - & - & General & - & - \\
        \midrule
        \textbf{HHMotion(Ours)} & \CheckmarkBold & \CheckmarkBold & \CheckmarkBold & \CheckmarkBold & \textbf{11} & \CheckmarkBold & \CheckmarkBold & \textbf{21.7h} & \textbf{Daily+Sports} & \textbf{15} & \textbf{1,000} \\

        \bottomrule[1pt]
        \end{tabular}
    }
    \vspace{-4mm}
        \label{tab:data_compare}
 \end{table*}

Inspired by the concept of the \textit{Turing Test}~\cite{turing2007computing} in artificial intelligence, we propose a \textbf{Motion Turing Test} for humanoid robots: \emph{given a robot’s motion, if a human evaluator cannot distinguish whether it originates from a human or a robot, the robot’s motion is considered to have passed the Motion Turing Test}. This test provides an intuitive and motion-centered approach to evaluating the human-likeness of robot motions, offering valuable insights for optimizing motion generation. Similar thinking about \textit{Turing Test} has been proposed previously~\cite{brooks1991intelligence, meyer1991proceedings, pfeifer2001understanding, pfeifer2006body, ortiz2016we}. For example, the Embodied Turing Test~\cite{zador2023catalyzing} evaluates whether an AI animal model is indistinguishable from its living counterpart. Different from them, we specifically focus on the motion of humanoid robots.

To enable this evaluation, we collected more than 21 hours of humanoid robot video data from major international events such as the World Robot Conference (WRC)~\cite{wrc2025}, World Artificial Intelligence Conference (WAIC)~\cite{waic2025}, World Humanoid Robot Games (WHRG)~\cite{whrg2025}, and others~\cite{lafan1_retarget_dataset}, which represent the current state of the art in humanoid robotics. However, as current robot appearances are still far from fully human-like, directly showing raw robot videos to humans would allow them to easily identify whether the motion is human or robotic based on appearance cues. To ensure that our Motion Turing Test focuses purely on motion rather than appearance, we estimate the corresponding SMPL-X~\cite{pavlakos2019expressive} models, a whole-body body model without texture information, from RGB videos of humanoid robots and humans.
We ask human evaluators to rate the human-likeness of these pose sequences. This process is shown in Fig.~\ref{fig:intro}.

Our collected dataset includes videos of 11 robot models (e.g., Unitree G1~\cite{unitreeg1}, ENGINEAI PM01~\cite{engineai_pm01}), covering a diverse set of \textbf{15 motion categories}, including standing, walking, running, and other daily and sports activities. We recruited 10 human participants performing the same category actions to enable direct \textbf{H}uman–\textbf{H}umanoid \textbf{Motion} comparison. Notably, we also collected motion data from simulated robotic environments and invited humans to mimic specific robot movements. The proposed \textbf{HHMotion} contains \textbf{1,000} video clips of humans and humanoids. % the distribution of the dataset statistics is depicted in Fig.~\ref{dataset}.

This work involved \textbf{30 human annotators} to score all pose sequences on a \textit{Likert scale}~\cite{likert1932technique} of 0 to 5, where 0 denotes ``completely robotic'', and 5 denotes ``completely human-like''. On average, each participant spent approximately one minute per video, totaling over \textbf{500 hours} of annotation effort for the entire dataset. We find that, despite significant advancements in humanoid robot motion, contrary to common misconceptions, existing robot poses still exhibit noticeable differences from human poses that humans can relatively easily distinguish. In particular, motions such as boxing received noticeably lower scores, while actions like walking show a closer resemblance to human ones. Tab.~\ref{tab:data_compare} summaries the difference among HHMotion and other datasets. \emph{As far as we know, this is the first dataset that rates the human-likeness of motions.} 

%Building upon this, we organized the collected data to propose a new benchmark \textbf{HHMotion} containing human and humanoid instances for assessing motion human-likeness in humanoid robots. This dataset enables a challenging pose-based human-likeness assessment task: given a 3D pose, predict how human-like quality it appears. To address this, we further propose a simple yet strong \textbf{baseline PTR-Net} that effectively learns to capture human-like motion characteristics and significantly outperforms existing large vision-language models (VLMs)~\cite{comanici2025gemini, achiam2023gpt, bai2023qwen}, even under well-designed prompting strategies. While humans can reliably distinguish between human and robotic poses, current VLMs still struggle with this fine-grained human-likeness perception. Our proposed baseline can also serve as a motion evaluation task for robot motion generation~\cite{allshire2025videomimic, tang2024humanmimic} and as a reward function in reinforcement learning~\cite{chen2024learning, gu2024humanoid} to improve humanoid motion synthesis.

%By Changwang
 The HHMotion enables a novel task that evaluates the human-likeness of motions. Given a 3D pose (either from a human or a humanoid), predict its degree of human-likeness ranging from 0 to 5. To establish a robust and trainable groundwork for the Motion Turing Test, we propose a simple baseline, Pose-Temporal Regression Network (\textbf{PTR-Net}), framing the evaluation of motion human-likeness as a quantitative regression task. PTR-Net notably surpasses several large vision-language model (VLM)-based methods ~\cite{comanici2025gemini, achiam2023gpt, bai2023qwen}, even with well-crafted prompting strategies. Our proposed baseline can also serve as a motion evaluation task for humanoid robot motion generation~\cite{allshire2025videomimic, tang2024humanmimic} and as a reward model in reinforcement learning~\cite {chen2024learning, gu2024humanoid} to enhance humanoid robot motion synthesis.

%While humans can accurately differentiate between human and robotic poses, current VLMs still encounter challenges in discerning fine-grained human-likeness perception.

Both the dataset and baseline implementations will be open-sourced to benefit the research community.

%-----------version 2
%现有人形机器人已经取得了非常大的进展，已经能够做出动作非常逼真的动作。在最近的机器人大会中，机器人能够非常逼真的行走、走路，挥手，甚至跳舞、体操。鉴于人形机器人在动作方面的巨大进展，借鉴于人工智能的图灵测试，我们提出一个人形机器人动作姿态的图灵测试问题：给定一个机器人的pose，如果一个人类无法区分这个pose是机器人还是人类的pose，那么就认为人形机器人的动作通过了机器人动作图灵测试。这个问题可以用于评估机器人动作的拟人程度，可以用于机器人动作生成的优化。

%为此，我们采集了来自 WRC 等国际顶级机器人赛事的 xxx 小时人形机器人视频数据，这些视频代表了当前人形机器人动作生成的最高水平。然而，由于现阶段的机器人外观尚未完全类人，若直接展示原始视频，人类评估者可以轻易地通过外观判断出其是否为机器人。为了让评估聚焦于“动作姿态”本身，我们采用基于 RGB 视频估计 SMPL-X 模型 的方式，将机器人和人类的动作统一映射为三维姿态，再由人类对这些姿态的“类人度”进行评分。

% 我们的数据集涵盖多种机器人型号（如 xxx、xxx），包含站立、行走、跑步等多种动作类型。我们同时邀请人类志愿者完成相同的动作，以实现人类与机器人的直接对比。此外，还包括仿真环境中的机器人动作数据，并邀请人类模仿机器人执行指定动作。整体数据统计如图xxx 所示。

% 本研究共邀请了 xxx 位人类评估者 对所有姿态视频进行 0–5 分打分，其中 0 表示“完全不像人类”，5 表示“完全像人类”。每位参与者平均花费 xxx 小时 进行评分，总计标注时间约 xxx 小时。结果显示，尽管人形机器人动作质量已有显著提升，但与common mis-understanding普遍认知不同，其动作姿态与人类仍存在明显差距，人类可以相对较为容易区分出来。尤其是在 xxx 动作上，得分显著偏低；而在 xxx 动作上，机器人的表现则更接近人类。

% 这个工作我们将采集的数据进行整理提出了一个新的数据集，这个数据集包含了xxx,xxx的数据。这个数据集可以enable这么一个机器人动作拟人度的task，给定一个pose，这个task告诉我们这个动作的拟人度。这个task是比较challenging的，尽管人类已经能够较好的区分机器人的动作和人类的动作，我们发现采用现有大模型的方式是无法有效的判别拟人度的，即使是通过精心设计的LLM方式还是存在有较大的误差。我们提出了一个简单的baseline，这个baseline表现比基于大模型的方式显著的好。这个baseline可以用于评估生成的机器人动作的拟人程度，从而用于指导机器人动作的生成，比如作为机器人动作生成强化学习的奖励函数。这个工作的数据集和baseline都将开源，to benenfit the community. 

%% file: sec/related.tex
\section{Related Work}
\label{sec:related}

\subsection{Human and Humanoid Robot Pose Estimation}

Research on 3D Human Pose Estimation (HPE) commonly employs parametric body models such as Skinned Multi-Person Linear Model (SMPL)~\cite{SMPL2015} and Skinned Multi-Person Linear Model-X (SMPL-X)~\cite{pavlakos2019expressive} to reconstruct 3D human pose and shape from 2D visual inputs. RGB-based HPE methods~\cite{Bogo2016KeepIS, chen2021sportscap, PIFU_2019ICCV, GFPose2023, TORE2023, VIBE, PARE_ICCV2021, pifuhd, TailorNet_CVPR2020, li2022cliff, MAED, BEV, ROMP, Su2020RobustFusionHV, MotionBert2023, yang2021s3, wang2023nemo, ICON} have achieved remarkable progress. Recently, world-grounded RGB-based HPE approaches~\cite{shen2024world, shin2024wham, yan2024reli11d, yan2023cimi4d, yan2025climbingcap, li2025coin, kocabas2024pace, sun2023trace, ye2023decoupling, yuan2022glamr} recover human motion in real-world coordinates, enabling temporally coherent, metrically aligned 3D reconstructions.

%Methods such as VIBE~\cite{VIBE} and PARE~\cite{PARE_ICCV2021} employ temporal modeling or refinement modules to ensure smooth pose sequences. Transformer-based approaches like BEV~\cite{BEV} and MotionBERT~\cite{MotionBert2023} further capture long-term motion dependencies through spatiotemporal representations. 
%Meanwhile, methods such as CLIFF~\cite{li2022cliff} and MAED~\cite{MAED} jointly estimate camera parameters and global body orientation, achieving accurate 3D localization in complex scenes.
%GVHMR~\cite{shen2024world} introduces a world-grounded formulation that reconstructs absolute 3D trajectories that global consistency enables a unified representation across subjects and views, making it particularly suitable for evaluating human and humanoid motions within a shared kinematic space.

Research on humanoid robot pose estimation has primarily focused on 2D pose estimation~\cite{heo2024diverse, amini2021real, ulya2023hiroposeestimation}. In recent years, 3D pose estimation for humanoid robots~\cite{cho2024full}  has also emerged. However, they are limited to specific types of robots, which limits their applicability.

%Research on humanoid robot pose estimation has primarily focused on 2D pose estimation~\cite{heo2024diverse, amini2021real, ulya2023hiroposeestimation}, which aims to accurately predict the 2D coordinates of humanoid robot joints from image data. However, such 2D representations limit their applicability in robotic tasks that require 3D interaction with the environment. In recent years, 3D pose estimation for humanoid robots~\cite{cho2024full}  has also emerged. However, they are limited to specific types of robots, which limits their applicability.

% 人形机器人姿态估计技术的研究主要集中于二维姿态估计方向[]，该方向旨在从图像数据中精确预测人形机器人关节点的二维坐标。然而，二维表示限制了其在需要对环境进行三维交互的机器人任务中的应用。近年来，学术界也出现了针对人形机器人的三维姿态估计研究[]。

\subsection{Human and Humanoid Motion Generation}

Human motion generation aims to synthesize natural and diverse motion sequences for both virtual humans and humanoid robots~\cite{zhu2023human}. Recent methods span autoregressive architectures~\cite{posegpt, motiongpt, T2M-GPT}, diffusion-based generative models~\cite{tevet2022human, liang2024intergen, shafir2024human, xie2024omnicontrol, bodiffusion}, and adversarial frameworks~\cite{barsoum2018hp, malek2023adversarial}. Multimodal extensions integrating text or audio guidance further enhance motion fidelity and expressiveness~\cite{zhang2024motiondiffuse, zhang2023generating}. 

In humanoid robotics, imitation learning approaches~\cite{allshire2025videomimic, tang2024humanmimic, okami2024, fu2024humanplus} and reinforcement learning with physics-based control~\cite{chen2024learning, gu2024humanoid, CassieJumpLi2023, radosavovic2024real, fan2025one, he2025hover} have enabled robots to perform human-like tasks. Motion retargeting techniques~\cite{baek2024human, mao2024learning, he2024learning, he2024omnih2o, yang2025omniretarget, harmon2024, tao2021visual} further bridge human–robot embodiment gaps. However, despite rapid progress in motion generation, a unified and quantitative evaluation of motion human-likeness remains largely unexplored.

\subsection{Motion Datasets}
Despite recent advances in human and humanoid motion generation, a standardized dataset for evaluating the human-likeness of motion remains absent. Existing motion datasets~\cite{AMASS_ICCV2019, Ionescu14a, lidarcap, lin2023motion} focus solely on humans rather than humanoid robots. With the rise of motion retargeting methods that transfer human movements to humanoid robots~\cite{he2024omnih2o, yang2025omniretarget, baek2024human}, many robots can now mimic human actions. The LAFAN1 Retargeting Dataset~\cite{lafan1_retarget_dataset} aims to make humanoid robot motions more natural. It is based on the original LAFAN1~\cite{harvey2020robust} motion capture dataset, where human motion data is retargeted to a humanoid robot~\cite{unitreeg1}. However, these approaches often exhibit a noticeable gap between simulation and real-world performance. 
%Table~\ref{tab:data_compare} summary the difference among HHMotion and others.

%Existing human motion datasets (e.g., AMASS~\cite{AMASS_ICCV2019}, Human3.6M~\cite{Ionescu14a}) primarily evaluate pose reconstruction or generation accuracy, focusing solely on human subjects and lacking cross-domain comparability with humanoid robots.

% Additionally, due to the emergence of methods for retargeting human motions to robots in the field of robotics~\cite{he2024omnih2o, yang2025omniretarget, baek2024human}, an increasing number of humanoid robots directly mimic human actions to enhance their movement capabilities. However, a significant issue with such approaches is that while robotic motions achieve satisfactory performance in simulation environments, the transition from simulation to reality faces substantial challenges.

%% file: sec/dataset.tex
\section{Human-humanoid Motion Dataset}
\label{sec:dataset}
% In this section, we introduce the main components of HHBench dataset. Sec~\ref{sec:3.1} reviews existing benchmarks and motivates the need for human-likeness evaluation across humans and humanoids. Sec~\ref{sec:data_construct} describes the construction of HHBench, while Sec~\ref{sec:humanscore} details the human preference annotation process. The experimental evaluations and insights derived from HHBench are discussed in Sec~\ref{benchmark} and~\ref{sec:exp}.
\begin{figure}[t]
    \centering
    \includegraphics[width=1.0\linewidth]{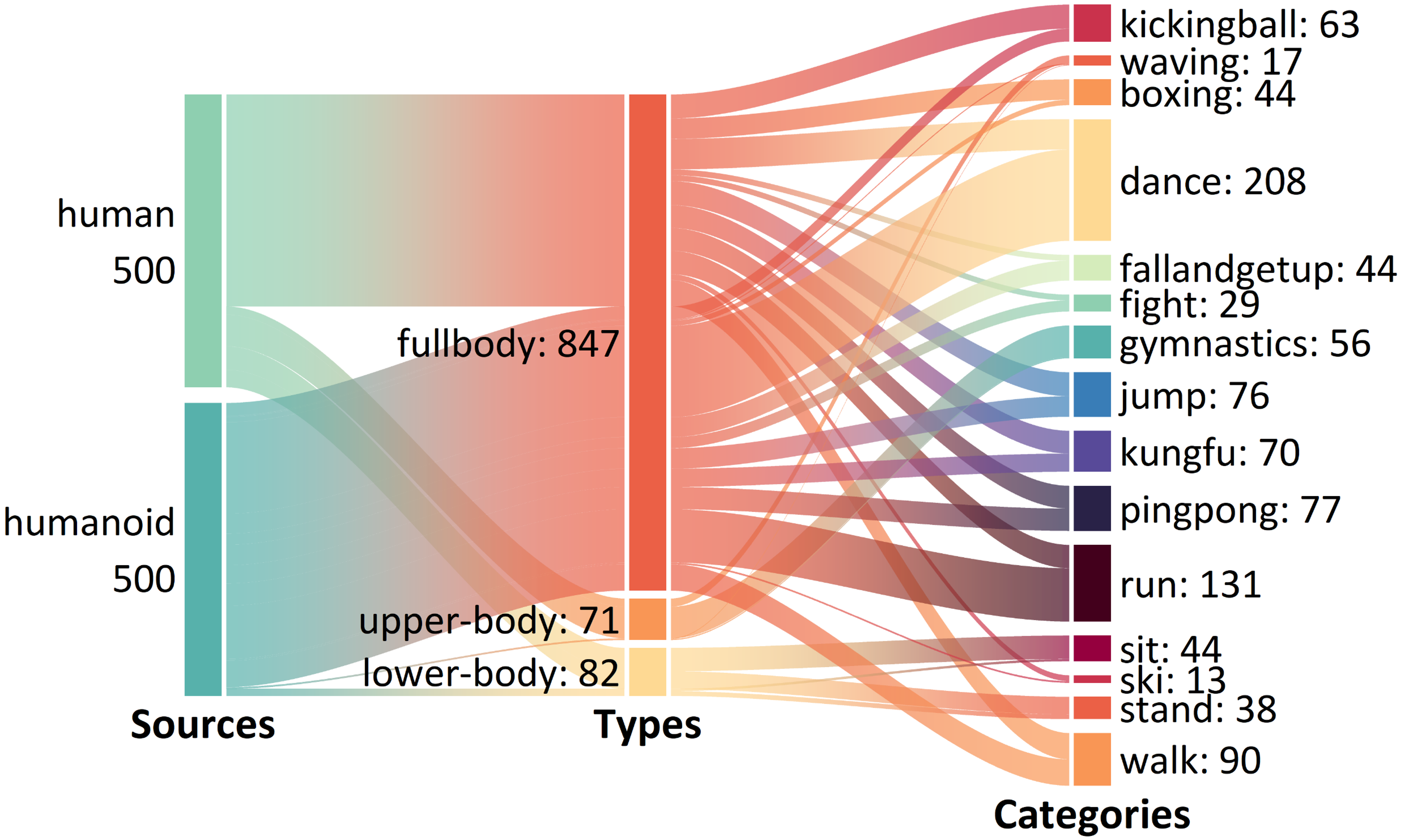}
    % \vspace{-2mm}
    \caption{\textbf{Action sources, types, and category distribution} in the HHMotion dataset, illustrating the diverse actions of both humans and humanoid robots.}
    \label{dataset}
    \vspace{-5mm}
\end{figure}
%\subsection{Benchmarks for Humanoid Motion}\label{sec:3.1}

\subsection{Dataset Overview}
\label{sec:3.1}
% Despite rapid advances in human and humanoid motion generation and control, a systematic benchmark for evaluating human-likeness in motion is still lacking. Existing human motion benchmarks (e.g., AMASS~\cite{AMASS_ICCV2019}, Human3.6M~\cite{Ionescu14a}) primarily focus on pose reconstruction or generation quality, targeting only humans and failing to enable cross-domain comparisons between humans and robots.

% Current benchmarks for humanoid motion quality mostly emphasize objective metrics such as task completion, efficiency, robustness, and end-effector trajectory accuracy~\cite{farhat2024walking, allshire2025videomimic, Huang2023quasi, liu2025opt2skill, rutili2024humanoidmotionsurvey, Wang2024reinforcement, meixner2024towards, yang2025omniretarget}. However, even when these metrics are high, the resulting robot motions may still lack the naturalness and fluidity perceived by human observers. In human-robot interaction scenarios, humans’ subjective perception of robot behavior is equally critical~\cite{gaebert2025effects, kunold2023not, cardenas2024evaluation}. Existing evaluation systems generally overlook the measurement of human-likeness in motion, including aspects such as motion quality, aesthetic appeal, and anthropomorphism.

% To address this gap, we propose xxBench, a unified evaluation framework for both humans and humanoid robots. By constructing a shared human-robot motion representation space, xxBench enables cross-domain and fair assessment.

\begin{figure*}[h]
    \vspace{-2mm}
    \centering
    \includegraphics[width=1.0\linewidth]{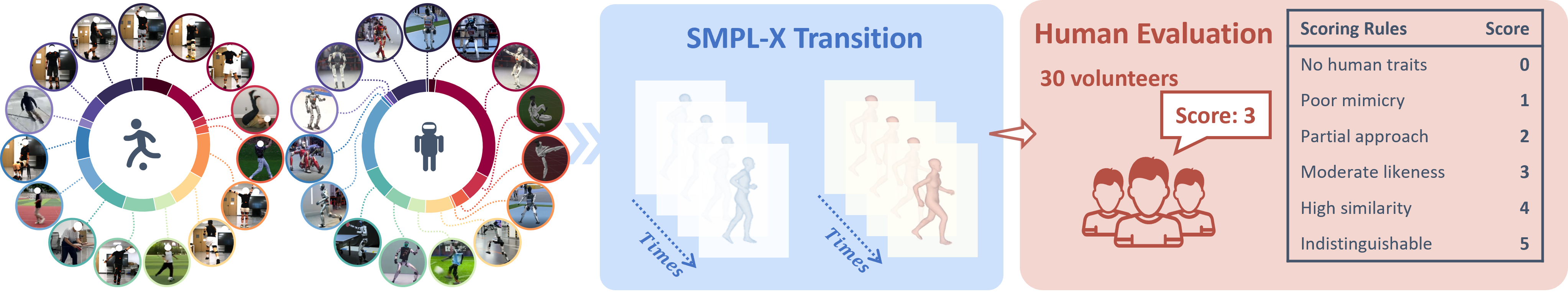}
    \caption{\textbf{Overview of the human scoring pipeline}, where all the humanoid robot and human motions are converted into SMPL-X poses and evaluated by human annotators. The resulting 0–5 scores quantitatively assess the human-likeness of each motion.}
    \label{fig:score_pipeline}
    \vspace{-4mm}
\end{figure*}

In this work, we propose the Human-Humanoid Motion (HHMotion) Dataset. Overall, HHMotion contains a balanced and diverse set of humanoid robot and human motion sequences, totaling \textbf{21.7h} of raw video data across both real-world and simulated environments. It consists of motions from \textbf{5 different sources}: real robot motions from significant events, robot motions from simulation, human motions from volunteers, human motions from volunteers mimicking robots, and human motions from YouTube. The dataset covers a wide range of \textbf{15 action categories} and \textbf{11 humanoid robot} models. The action category distribution is depicted in Fig.~\ref{dataset}.

HHMotion consists of \textbf{10 subjects} performing actions in the same category as robots. Moreover, to increase the dataset's diversity, we collected multiple human motion videos from YouTube (see detailed dataset sources and humanoid models in Supplementary File).

Each video is standardized into a 5-second clip, resulting in \textbf{1,000 motion clips}. Each motion clip is scored by human annotators on a Likert scale of 0 to 5. 0 represents ``complete robotic," whereas 5 represents ``complete human-like" (which is described in Sec.~\ref{sec:humanscore}).

%The comparison of HHMotion to existing human-humanoid motion datasets is summarized in Tab.~\ref{tab:data_compare}. 

%To fill this gap, we introduce \textbf{HHBench}, a unified benchmark for assessing motion human-likeness across both humans and humanoid robots. By establishing a shared motion representation space, HHBench enables quantitatively fair evaluation of motion quality.

% 尽管人类与人形机器人运动生成与控制技术发展迅速，但目前仍缺乏系统化的huamanoid运动质量评估基准。现有的人体运动基准（如 AMASS[]、Human3.6M[]）主要关注人体姿态重建或生成质量评估，评测对象仅限于人类本身，未能实现人与机器人之间的跨域比较。
% 当前针对人形机器人运动质量的评估基准，大多聚焦于任务完成率、效率、鲁棒性以及末端执行器轨迹精度等客观指标【…】。然而，即便这些指标表现优异，机器人所呈现的运动仍可能缺乏人类观察者所感知的自然性与流畅度。在人机协作场景中，人类对机器人行为的主观感受同样至关重要【…】。现有评价体系普遍忽视了对运动行为“类人程度”的衡量，包括其运动质量、美学表现与拟人化水平。为应对上述空白，我们提出了一个面向人类与人形机器人的统一评测框架xxBench，建立了人机统一运动表征空间，实现了跨域、公平的评估。

\subsection{Dataset Construction}
\label{sec:data_construct}

%One question that guides this benchmark is: How can the human-like behaviors of humanoid robots be evaluated? We consider that a corresponding standardized dataset for both humans and humanoid robots is required. Although datasets for humans and humanoid robots have been established separately, there remains a shortage of motion-oriented datasets with unified action categories for both, which is crucial for both humanoid robotics and human motion research. Based on this, we have constructed HHMotion, a standardized human and humanoid robot motion dataset.

%引导这个benchmark构建一个重要的问题就是：我们如何才能够对人形机器人的类人程度进行评价？我们认为这一个对应的标准化人类和人形机器人数据集是需要的。尽管已经分别存在人类和人形机器人数据集，但是对于这二者具有统一动作类别的motion指向的数据集仍然紧缺，而这对于人形机器人和人类运动领域都是至关重要的。基于此，我们构建了一个标准化的人类和人形机器人运动数据集XXDataset.
\subsubsection{Humanoid robot data}
To uncover how to establish the world's most advanced motion benchmark for humanoid robots, we collected 21.7 hours of original videos from the WRC~\cite{wrc2025}, WAIC~\cite{waic2025}, and WHRG~\cite{whrg2025}. Data from the WRC and WAIC were sourced from platforms like YouTube, while 2025 WHRG data originated from official competition broadcasts. This authentic humanoid robot dataset represents the world's most advanced level of humanoid robotics development. Among these conferences, we collected data from 11 robot models, encompassing algorithms from 28 different teams. %These real humanoid robot actions can be categorized into 15 action categories as shown in Fig.~\ref{dataset}.

%The LAFAN1 Retargeting Dataset~\cite{lafan1_retarget_dataset} aims to make humanoid robot motions more natural. It is based on the original LAFAN1~\cite{harvey2020robust} motion capture dataset, where human motion data is retargeted onto Unitree's humanoid robot~\cite{unitreeg1}.

Many robot motion generation methods adopt sim-to-real approaches, which rely on simulated robot motions. To comprehensively evaluate the human-like scale of humanoid robot motions, we also collected robot motion sequences in simulated environments. We visualized and recorded the LAFAN1 Retargeting Dataset~\cite{lafan1_retarget_dataset} in a simulation environment, collecting raw robotic motion sequences across 7 action categories: dance, fall and get up, fight, jump, kicking ball, run, and walk.

% After a rigorous data sanitation and standardization process, we retained about 5,000 seconds of high-quality motion clips (including human and humanoid). Low-resolution, occluded, or truncated videos were first removed, and all remaining sequences were trimmed into 5-second segments containing complete action sequences.
To enable more precise evaluation for each action category, each video action in the dataset was segmented into 5-second action clips, each containing a simple but complete action. 
In total, we obtained \textbf{500 humanoid motion clips}, comprising 257 clips from real humanoid robots and 243 clips from robots in the simulation environment.
After a rigorous data sanitation and standardization process, we retained about 5,000 seconds of high-quality motion clips (including human and humanoid). Low-resolution, occluded, or truncated videos were removed.

%为了发掘如何才能获得世界上最先进的人形机器人的运动水平基准，我们收集了XX小时的2025 world robot conference （2025 WRC）[]、2025 world artificial intelligence conference （2025waic）[] 和2025 world humanoid robot games （2025WHRG）[]的原始视频序列，其中2025 WRC和2025waic的数据为YouTube等网站搜取，2025WHRG的数据为官方比赛转播画面，这些真实的人形机器人数据代表了人形机器人世界最先进发展水平。在这两个会议上我们共收集了来自xx种机器人型号，分别是：...,其中包括了xx种不同团队的算法。这些真实的人形机器人动作可以被分为15个动作类别，包括跑步，跳舞....。
%除此以外，由于在机器人领域中从人类动作重定向到机器人动作方法的涌现，越来越多的人形机器人直接mimic人类的动作以提升机器人的运动能力。然而这类方法存在的一个重要问题是：机器人的动作在仿真环境下达到较好的水平，但sim-to-real的过程却受到了巨大的阻力。因此，为了更加全面的评估人形机器人的动作像人程度，我们还收集了仿真环境下的机器人运动序列。LAFAN1 Retargeting Dataset【】 是一个旨在使类人机器人运动更自然的数据集，它基于原始的 LAFAN1【】运动捕捉数据集，将其中的人体运动数据重定向到Unitree的类人机器人上。我们将LAFAN1 Retargeting Dataset在仿真环境下进行可视化并进行录制，共收集了xx小时的原始仿真环境中机器人的运动序列，包含了7种动作类别，涵盖了dance、fallandgetup、fight、jump、kicking ball、run、walk。
%为了更好的为每个类别的动作进行精细化评估，数据集中的每个视频动作都被划分成了一个5秒钟的动作剪辑,该剪辑中包含一个简单但又完整的动作。最终，我们获得了500条人形机器人的动作clip,其中包括257条真实人形机器人和243条仿真环境下的机器人。

\subsubsection{Human motion data}
\label{sec:humandata}
To enable comparative evaluation between robotic and human motions, we collected corresponding human actions based on the categories gathered from humanoid robots. For each category, 10 subjects performed 25–30 specific motions. 
% , resulting in xx hours of video sequences. 
This dataset comprehensively covers all robotic motion categories collected, spanning both indoor and outdoor scenarios. 
To align with robotic motions, the collected data was consistently edited into 5-second clips. 
In addition, in order to increase the diversity of the human sample, videos of various people performing the corresponding action categories were collected from YouTube. In total, we collected \textbf{500 human motion clips} across 15 categories, comprising 365 clips from 10 subjects and 135 clips from the internet.

%除此以外 我们还增加了一些人类模仿机器人的数据
Notably, to reinforce the Motion Turing Test principle of our benchmark, we also included a subset where human subjects \textbf{intentionally imitated} the movement patterns of humanoid robots (refer to the Supplementary File for the detailed evaluation). The goal was to create ambiguous motion samples that blur the distinction between human and robotic behaviors, thereby making the evaluation more challenging and faithful to the essence of Motion Turing Test.

% Finally, all collected motions were categorized into 15 motion types, which were further grouped according to human body joint into full-body, upper-body, and lower-body movements. Among them, full-body movements constitute xx percent of the dataset, demonstrating that the difficulty level of actions in our dataset presents serious obstacles to humanoid robots. This standardized human–humanoid dataset serves as the foundation for evaluating the degree of human-likeness in humanoid robot motion.
% 为了能够对比评估机器人与人的动作，我们根据收集到的人形机器人的动作类别采集相同的人类动作，每种动作各采集10名志愿者，每名志愿者执行了25-30个细分的动作，总共采集了xx小时的视频，其涵盖了上述收集的机器人运动中的所有动作类别，包括室内和室外的场景。为了和机器人的运动对齐，我们将采集到的数据进行一致性的剪辑为5秒的clip。除此以外，为了增加人类样本的丰富度，还从YouTube上收集了139条相应动作类别的各类人群的视频，并对运动进行剪辑。总的来讲我们共收集了500条人类的动作片段，其中包括261条采集的10名志愿者的数据，139条互联网上的运动视频。
%重要的是，为了强化我们基准测试中的图灵测试原则，我们还纳入了一个子集，其中人类参与者有意模仿了人形机器人动作模式。其目的是创建模糊人类与机器人行为区别的模糊动作样本，从而使评估更具挑战性，并更忠实于运动图灵测试的本质。
%我们将15种动作类别按照人类的关节进行划分，分成了full-body、上半身和下半身的运动，其中全身的运动占数据集的百分之xx，这也证明了我们数据集中的动作难度对于人形机器人具有一定的挑战性。以上人类-人形机器人标准化数据集的构建为评估机器人像人的程度提供了必要的组成。

\subsection{Human-humanoid Pose Estimation}
\label{hpe}
%强调使用gvhmr的严谨性
As humanoid robots still exhibit distinctive appearances such as metallic shells and exposed joints, evaluators may rely on visual cues rather than motion to distinguish them from humans. To ensure that our Motion Turing Test focuses purely on motion rather than appearance, we adopt a pose-level strategy that converts all videos into the SMPL-X~\cite{pavlakos2019expressive} model, ensuring that only motion information is focused for evaluation.

To obtain the SMPL-X model used for human-likeness evaluation, we systematically compared multiple SOTA human pose estimation (HPE)~\cite{shen2024world, shin2024wham, yan2024reli11d, yan2025climbingcap} and robot pose estimation (RPE)~\cite{amini2021real, heo2024diverse, cho2024full} methods. We found that existing robot-specific pose estimators often overfit to particular robot morphologies or kinematic constraints, leading to degraded generalization when applied to diverse humanoid models. In contrast, general HPE methods demonstrated stronger robustness and better alignment across human and humanoid subjects. Among these HPE methods, GVHMR~\cite{shen2024world} achieved the most stable and temporally coherent results in our experiments (refer to Supplementary File). Although it was initially designed for human motion, we observed that it generalized well to humanoid robots, providing accurate and temporally smooth reconstructions.  %This choice also enables a fairer comparison between human and robot motions.

\subsection{Human-likeness Scoring}
\label{sec:humanscore}
\begin{figure}
    \centering
    \includegraphics[width=1.0\linewidth]{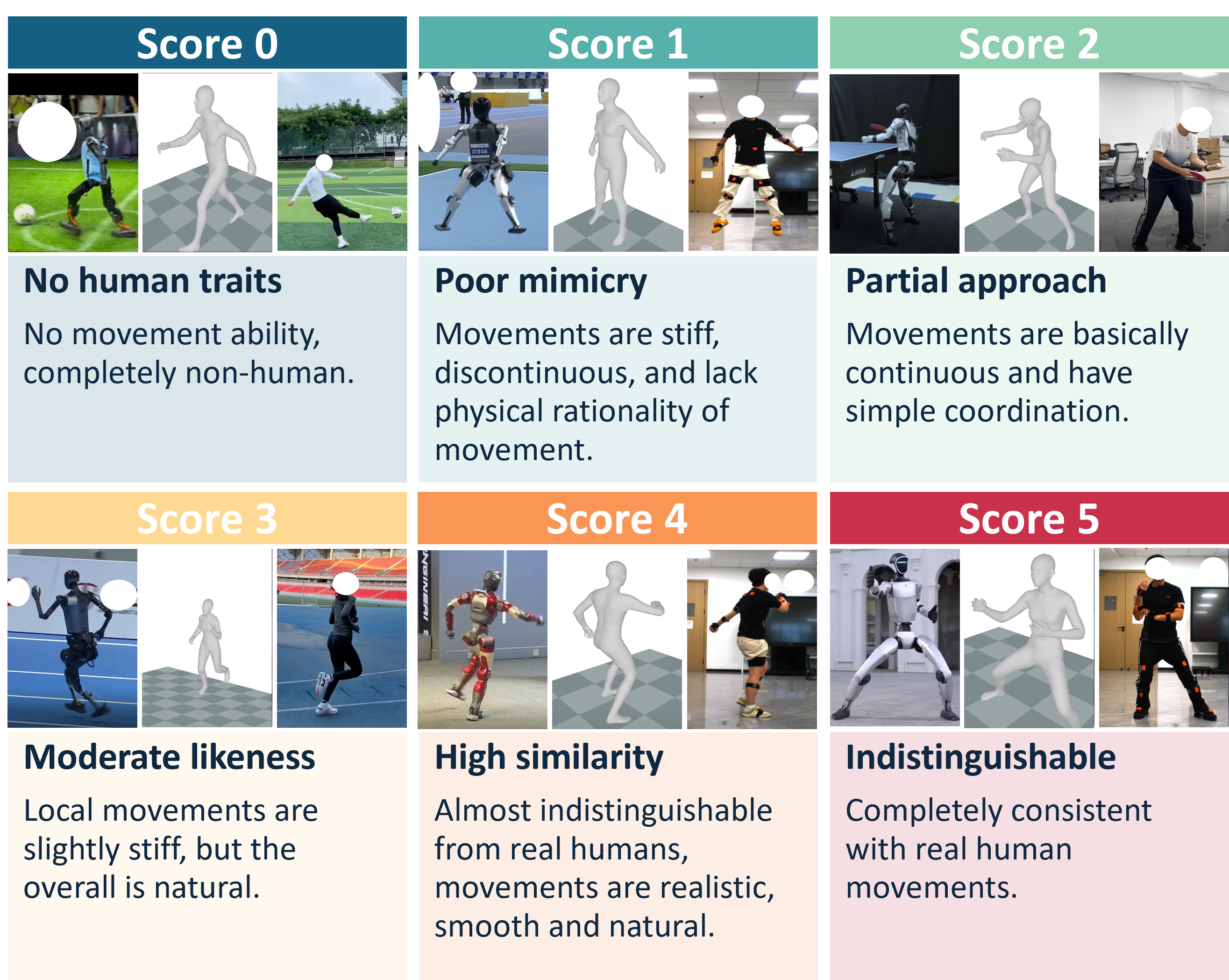}
    \vspace{-6mm}
    \caption{
    % Clear motion human-likeness scoring rules, employing integer scores from 0 to 5 to quantitatively assess each motion clip.
    \textbf{Human-likeness scoring rules} used in evaluating motion clips on a 0–5 Likert scale, focusing solely on motion quality.}
    \label{scoring}
   \vspace{-4mm}
\end{figure}

Current humanoid motion researches predominantly assess task-oriented metrics, such as completion rate, efficiency, robustness, or end-effector trajectory accuracy~\cite{farhat2024walking, allshire2025videomimic, Huang2023quasi, liu2025opt2skill, rutili2024humanoidmotionsurvey, Wang2024reinforcement, meixner2024towards, yang2025omniretarget}. However, high task performance does not necessarily imply perceptually human-like motion. In human–robot interaction, subjective impressions of naturalness, fluidity, and anthropomorphism are equally critical~\cite{gaebert2025effects, kunold2023not, cardenas2024evaluation}, yet these aspects are often overlooked in existing evaluation systems.

The human scoring pipeline is shown in Fig.~\ref{fig:score_pipeline}. To quantitatively assess the human-likeness of humanoid robot motions, we collected large-scale human annotations for each motion clip in the HHMotion. Specifically, we recruited \textbf{30 human evaluators} to assign human-likeness scores to all SMPL-X motion sequences on a 0–5 \textit{Likert scale}~\cite{likert1932technique}, where 0 indicates ``completely inhuman-like motion” and 5 indicates ``indistinguishable from human motion” as shown in Fig.~\ref{scoring}. Each participant was asked to carefully watch a set of motion sequences and evaluate how closely each motion resembled natural human movement in terms of posture, rhythm, and coordination. We have obtained IRB approval to conduct this study. All the volunteers and the annotators agree that their data can be used for research purposes.

For evaluation, we selected 500 human and 500 humanoid robot motion sequences, randomly mixed and shuffled to hide their sources, and presented them to each evaluator to prevent any prior bias toward a specific class or motion type. Each annotator evaluated 1000 sequences and spent approximately an average of 16.7 hours completing the task, leading to a total of over 500 annotation hours across all participants. Finally, we aggregated all annotators' scores to obtain the average human-likeness score for each action sequence, forming the “human evaluators annotation” within the HHMotion dataset.

%一致性检查的具体过程：
% To ensure the reliability and credibility of the collected human-likeness annotations, we conducted an Inter-Annotator Consistency Check across all participants. Specifically, we measured both the Mean Squared Error (MSE) of absolute deviations and the Spearman’s rank correlation coefficient between each annotator’s scores and the overall average scores. This dual-metric evaluation allowed us to detect annotators whose scoring patterns significantly deviated from the group consensus. As a result, five annotators were identified as inconsistent with the overall distribution and were excluded from the final dataset. The remaining 25 evaluators demonstrated high inter-annotator consistency and provided stable, accurate assessments. This rigorous filtering process ensures the credibility and robustness of the human preference annotations. Details are provided in Appendix 1.

%This annotated subset provides a benchmark for evaluating and training the anthropomorphic quality of generated robotic motions.

\begin{figure}[t]
    \centering
    \includegraphics[width=1.0\linewidth]{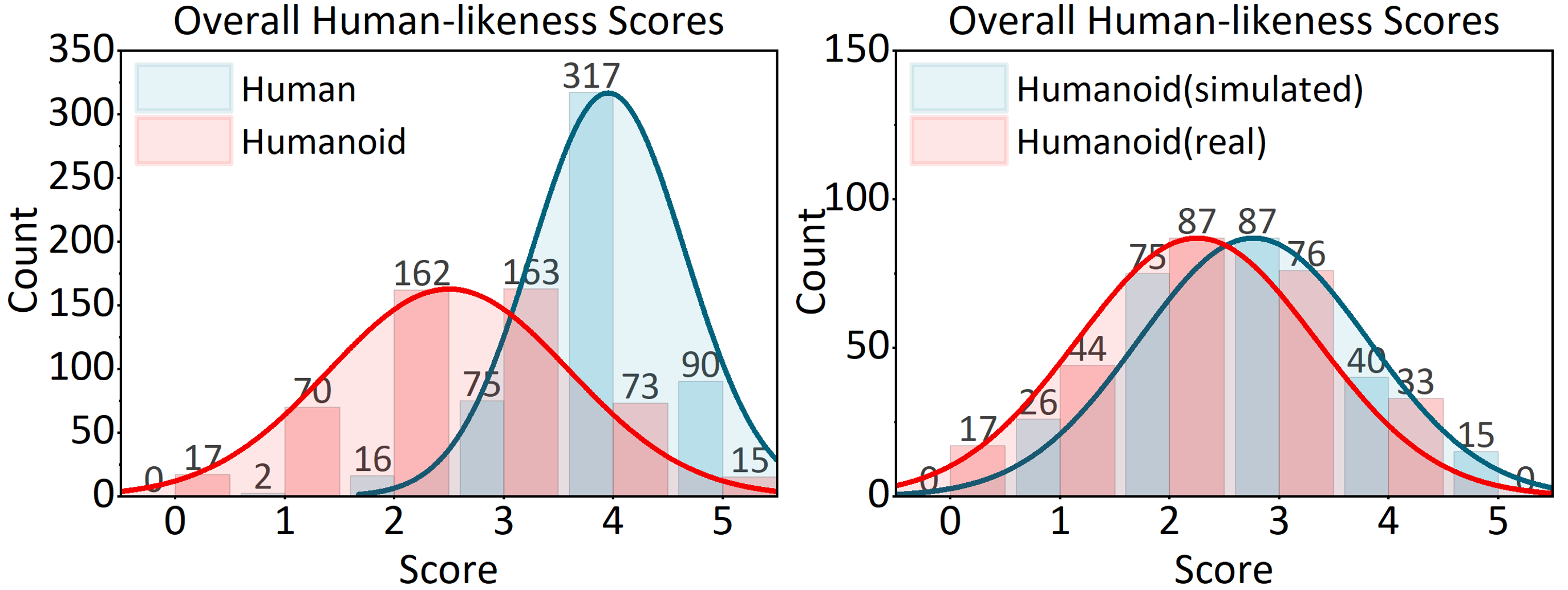}
    \vspace{-6mm}
    \caption{Overall distribution of motion human-likeness scores for \textbf{human and humanoid} motions (\textbf{left}) and human-likeness scores for \textbf{humanoid in simulation and real} scenarios (\textbf{right}).}
    \label{fig:overallscore}
    \vspace{-4mm}
\end{figure}

\subsection{Dataset Quality}
\textbf{Motion sequence Quality.} To ensure reliability of the obtained SMPL-X motion sequences for human and robot, all reconstructed motion sequences underwent manual cross-validation to remove failed or noisy samples (e.g., from occlusion or reflective surfaces). Furthermore, a subset of videos was manually annotated with bounding boxes for correction, where we also report detailed comparisons of tested HPE methods (refer to Supplementary File for details). 
% Some poorly estimated SMPL-X sequences are removed from the dataset. 
Through this pipeline, we obtain high-quality, temporally smooth SMPL-X motion sequences for both human and humanoid subjects, establishing a robust foundation for our Motion Turing Test benchmark.

\textbf{Annotation quality.}
An \textbf{I}nter-\textbf{A}nnotator \textbf{C}onsistency (IAC) check was calculated to confirm high agreement levels among human evaluators. As a result, five annotators were identified as inconsistent with the overall distribution and were excluded from the final dataset. The remaining 25 evaluators demonstrated high IAC and provided stable, accurate assessments. This rigorous filtering process ensures the credibility and robustness of the human-likeness score annotations (see IAC process and result in Supplementary).

\subsection{Analysis of Human-likness Scores}
To quantitatively understand how human evaluators assess the similarity between humanoid robot and human motions, we conducted a detailed analysis of the human evaluators' annotations across 15 action categories. Fig.~\ref{fig:overallscore} (left) illustrates the overall score distribution for human and humanoid motions, revealing that human actions consistently receive higher rating counts. Fig.~\ref{fig:overallscore} (right) illustrates the score distribution for humanoids in simulation and real scenarios, showing that humanoid actions in simulation environments perform better than real humanoid robots. 
This indicates that, despite recent advances in humanoid control, a noticeable gap in human-likeness remains between human and robotic motion quality.
Tab.~\ref{tab:score_diff} presents the top and bottom five categories ranked by the score differences between human and humanoid motions. In this comparison, our analysis focuses exclusively on real humanoid robots to ensure fairness and reliability.

%Further, such figure reveals that the simulated robot poses still significantly different from human motions.

% \begin{figure*}[h]
%     % \vspace{-2mm}
%     \centering
%     \includegraphics[width=0.9\linewidth]{figs/overview.png}
%     \caption{\textbf{Overview of the Motion Turing Test Benchmark.} Humanoid and human motions are converted into SMPL-X poses and evaluated by humans, supervised regressors, and VLM. The 0–5 scores quantify motion human-likeness.}
%     \label{fig:overview}
%     \vspace{-6mm}
% \end{figure*}

\begin{figure}[t]
    \centering
    % \vspace{-2mm}
    \includegraphics[width=1.0\linewidth, height=0.7\linewidth]{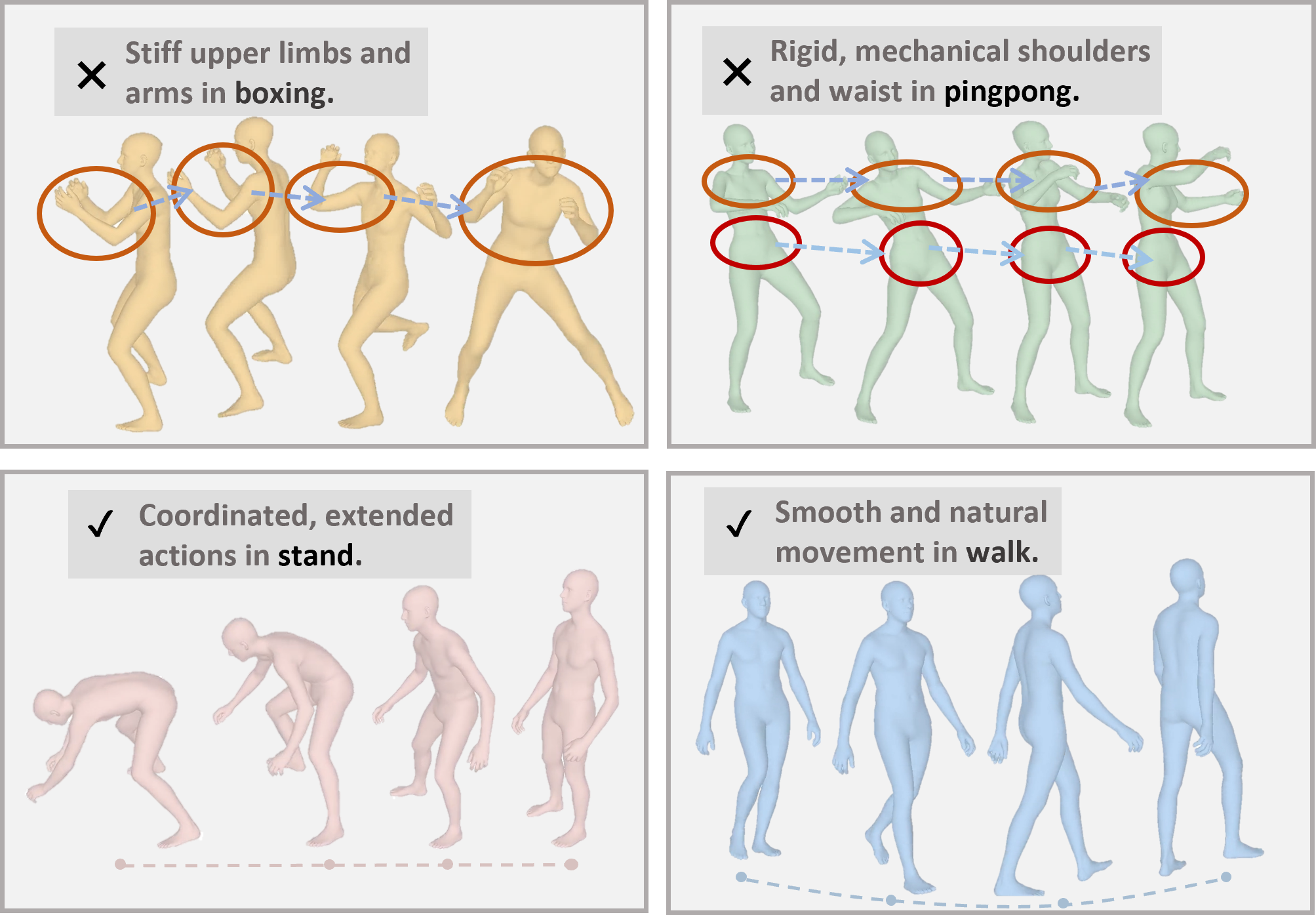}
    \vspace{-6mm}
    \caption{Representative SMPL-X sequences illustrate where humanoid robots perform poorly (upper part) and well (lower part).}
    \vspace{-4mm}
    \label{fig:badcase}
\end{figure}

Overall, our human-likeness score analysis demonstrates that, contrary to common misconceptions, there remains a significant gap between human and robotic movements, even with the remarkable advancements in humanoid robot motion control. Evaluators could easily distinguish humanoid motions from natural human movements, even under the appearance-disentangled evaluation setting. Notably, actions involving high-frequency coordination and rapid limb transitions, such as \textit{jump}, \textit{boxing}, and \textit{pingpong} (see upper part of Fig.~\ref{fig:badcase}), showed the largest score discrepancies, suggesting that humanoid robots continue to struggle with dynamic, contact-rich, and reactive movements.

\begin{table}[t]
\centering
% \vspace{-2mm}
\caption{\textbf{Top 5 and bottom 5 action categories} ranked by score difference between human and real humanoid motions. The scores here are calculated based on the average score of the last 25 annotators after the IAC check.}
\vspace{-3mm}
\renewcommand\arraystretch{0.8} 
\resizebox{\linewidth}{!}{
\begin{tabular}{lccc}
\toprule
\textbf{Action Category} & \textbf{Human} & \textbf{Humanoid} & \textbf{Score Difference} \\
\midrule
\multicolumn{4}{l}{\textit{Smallest Differences}} \\
stand  & 3.80 & 1.97 & 1.83 \\
sit    & 4.18 & 2.63 & 1.55 \\
ski    & 3.13 & 1.76 & 1.37 \\
walk   & 3.92 & 2.61 & 1.31 \\
dance  & 3.47 & 2.26 & 1.21 \\
\midrule
\multicolumn{4}{l}{\textit{Largest Differences}} \\
jump        & 4.43 & 1.20 & 3.23 \\
boxing      & 3.76 & 1.23 & 2.53 \\
run         & 3.73 & 1.47 & 2.26 \\
pingpong    & 4.33 & 2.09 & 2.24 \\
kicking ball & 3.93 & 1.79 & 2.14 \\
\bottomrule
\end{tabular}}
\label{tab:score_diff}
 \vspace{-6mm}
\end{table}

%Overall, despite the impressive progress in humanoid robot motion control, our human preference analysis shows that, in contrast to common misunderstandings, there is still a considerable gap between human and robotic movements. Even in the appearance-disentangled evaluation setting, evaluators were able to discern humanoid motions from actual human movements with ease. The biggest score gaps were found in activities like \textit{boxing}, \textit{pingpong}, and \textit{kung fu}, which require high-frequency coordination and quick limb changes. This suggests that humanoid robots still have difficulty with dynamic, contact-rich, and reactive motions.

In contrast, relatively smooth or cyclic motion types such as \textit{walk}, \textit{stand} (see lower part of Fig.~\ref{fig:badcase}), achieved closer human–humanoid score alignment. These findings indicate that current humanoid robots can reproduce structured or rhythmically repetitive movements with reasonable fidelity but still lack the fine-grained fluidity, adaptiveness, and balance control characteristic of human motor behavior.

Comprehensive score analysis for all \textbf{simulated} categories is provided in the Supplementary File for completeness. The analysis shows that the humanoid's performance in the simulated environment is superior to its performance in the real world.
%总的来说，我们的人类偏好分析表明，尽管仿人机器人运动控制取得了显著的进步，但人类和机器人的动作模式之间仍存在相当大的差距。评估者能够轻易地区分仿人的动作和自然的人类动作，即使是在隐匿外观的评估设置下也是如此。，涉及高频率的协调和快速肢体转换的动作，如拳击、乒乓球和功夫，显示出最大的得分差异，这表明仿人机器人在动态、接触丰富且反应性较强的运动方面仍存在困难。
%相比之下，相对平滑或循环的运动类型，如战斗、摔倒并起身以及踢球，实现了更接近的人机得分一致性。这些发现表明，当前的仿人能够以合理的精度重现结构化或有节奏重复的动作，但仍缺乏人类运动行为所具有的精细流畅性、适应性和平衡控制特性。

%\subsection{Where Humanoid Robots Still Fail?}
%To illustrate where the humanoid robots still fail, we visualize four representative humanoid SMPL-X sequences as shown in Fig.~\ref{fig:badcase}. Boxing and pingpong show typical failure cases, which exhibit stiff upper limbs, rigid shoulders, and mechanical waist rotations, leading to unnatural and discontinuous movements. In contrast, kicking a ball and walking motions are successes in coordinated joint motion, extended actions, and smooth transitions that appear closer to human performance.

%The results confirm that while humanoid robots can mimic the general structure of human motion, they still lack the inter-joint coordination and temporal fluency that define natural human movement. This underscores the importance of our benchmark in quantitatively evaluating and guiding future improvements in human-like motion generation.

%% file: sec/benchmark.tex
\section{Robot Motion Turing Test Benchmark}
\label{benchmark}
To quantitatively assess the human-likeness of humanoid robot motions, we construct a unified \textbf{Robot Motion Turing Test Benchmark} based on the HHMotion. 
%This benchmark aims to evaluate whether LLMs or motion pose assessment networks can approximate human perception of motion realism. The evaluation includes three key components: 1) task definition and benchmark formulation; 2) LLM-based automatic evaluation; 3) supervised xx baselines for human-likeness assessment.
This benchmark examines whether different models can approximate human judgments of motion human-likeness. We first describe the Motion human-likeness task (Sec.~\ref{sec:define}), then introduce our supervised PTR-Net for human-likeness prediction from pose sequences (Sec.~\ref{sec:network}) and describe the evaluation metrics for model–human alignment in Sec.~\ref{sec:metrics}.

%We further evaluate VLM as automatic motion assessors (Sec.~\ref{sec:llm}),
%为了定量评估人形机器人动作的类人程度，我们基于所收集的数据集和人类标注构建了一个统一的动作图灵测试基准。该基准旨在评估当前模型，无论是大模型还是动作评估网络能否接近人类对动作真实感的感知。评估包括三个关键部分：1）任务定义和基准制定；2）基于LLM的自动评估；3）用于评估类人程度的监督baseline。
\vspace{-2mm}
\subsection{Motion Human-likeness Assessment task}
\label{sec:define}
%In response to the central question proposed in the introduction, whether a given pose is more human-like or humanoid, we define a new evaluation protocol called the \textbf{Motion Turing Test}(MTT) task. 

The Robot Motion Turing Test is inspired by the classical \textit{Turing Test}~\cite{turing2007computing} but focuses exclusively on motion and pose realism, rather than visual appearance. Specifically, a humanoid robot is considered to ``pass” the Motion Turing Test if human evaluators cannot reliably distinguish whether a given pose sequence was performed by a human or a humanoid, based solely on its body motion information (without facial, textual, and color information). To implement such a Turing test, we propose a novel task for assessing motion human-likeness. We formulate the assessment of human-likeness as a quantitative regression task. 

In this task, a model is required to predict continuous human-likeness scores directly from motion sequences, aligned with human annotators' judgments. It takes a motion sequence as input and outputs a human-likeness score ranging from 0 to 5, where 0 represents a complete difference from a human and 5 indicates indistinguishability from a human. Each motion sequence is represented as a temporal series of the SMPL-X model. The sequence is normalized into a local root coordinate frame to remove global translation and rotation effects. 

%The background of the original RGB sequence is removed to avoid potential interfere.

%\textbf{Input}
%Each motion sequence is represented as a temporal series of 3D joint coordinates:
%\begin{equation}
%    \mathbf{X} = \{x_t \in \mathbb{R}^{J \times 3}\}_{t=1}^T
%\end{equation}
%where \(J\) denotes the number of joints and \(T\) the frame length. The sequence is normalized into a local root coordinate frame to remove global translation and rotation effects.

% To operationalize this concept, we leverage the prepared dataset described in Sec.\ref{sec:dataset}, which contains synchronized human and humanoid motion samples covering diverse daily and athletic activities. The task formulation includes two complementary evaluation pathways: supervised evaluation via PTR-Net (Sec.\ref{sec:network}) and LLM-based evaluation (Sec.\ref{sec:llm}).

%我们的数据集已经具备，针对回应我们introduction中提出的问题，如何判断是一个动作姿态是人形机器人或者是人，因此我们设计了两种，一种是llm评测的方式，一种是训练数据集的方式，用我们的模型去训练评估模型。
%针对引言中提出的中心问题，给定的姿势是来自人类还是类人机器人，我们定义了一个新的评估protocol，称为运动图灵测试任务。该任务受经典图灵测试的启发，但仅专注于运动和姿势的像人感，而非视觉外观。具体而言，如果人类评估者仅基于SMPL-X 身体表示无法可靠地区分给定的姿势或运动序列是由人类还是机器人完成的，则认为人形机器人通过了运动图灵测试。为了将这一概念付诸实践，我们利用了sec 3节中描述的精心策划的数据集，该数据集包含同步的人类和人形机器人运动样本，涵盖了各种日常和体育活动。任务的制定包括两个互补的评估途径：1. 基于大语言模型的评估 sec4.2。2. 通过训练模型进行监督评估sec4.3。
\vspace{-1mm}
\subsection{PTR-Net: A Simple Baseline }
\label{sec:network}
\vspace{-1mm}
%To establish a solid and learnable foundation for the MTT, we propose a trainable baseline Pose-Temporal Regression Network (PTR-Net),  The designed PTR-Net is a lightweight yet expressive model unlike general motion representation models, and PTR-Net is trained from scratch on human preference annotations.

\begin{figure}[t]
    \centering
    \vspace{-2mm}
    \includegraphics[width=1\linewidth]{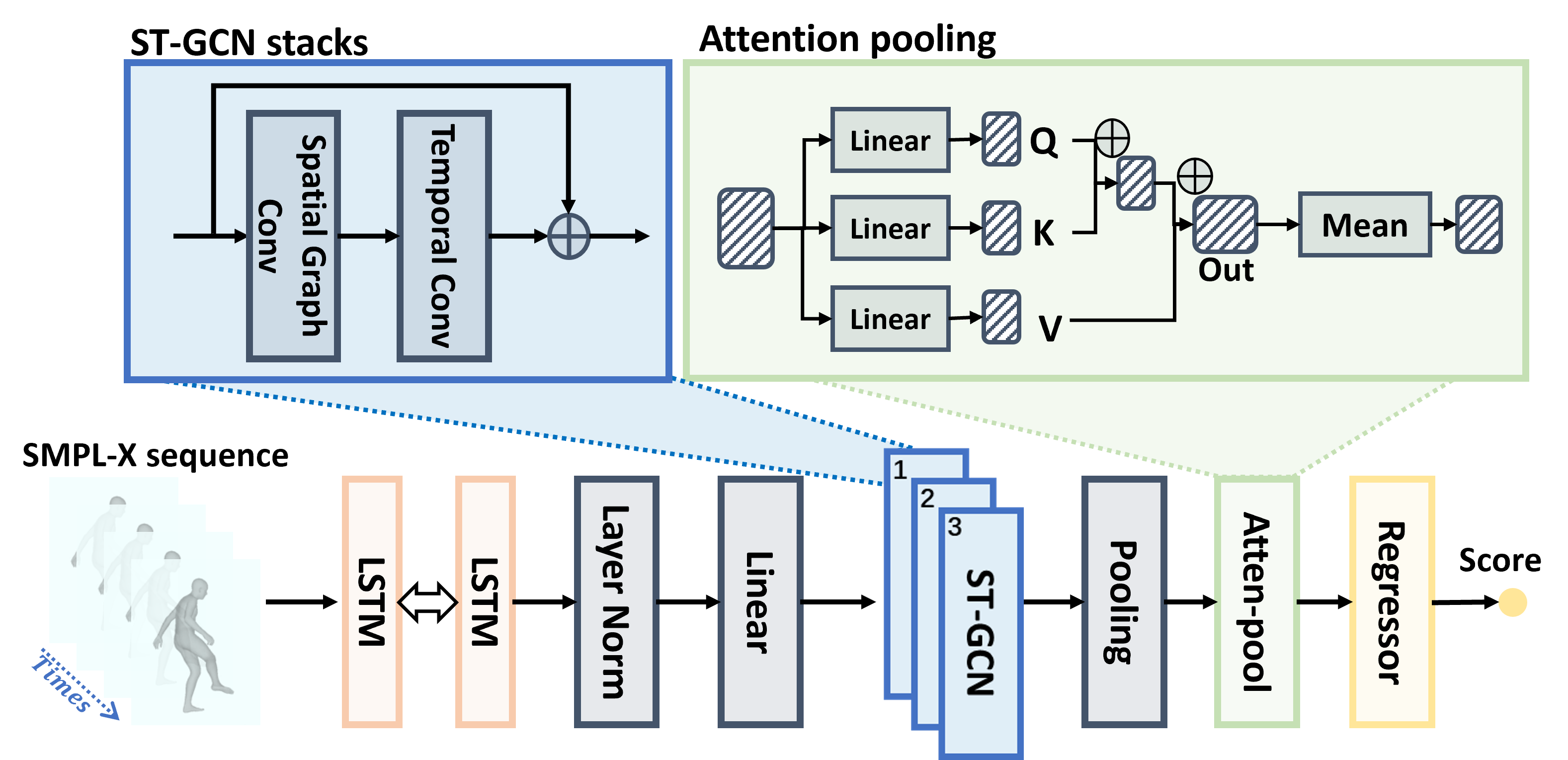}
    \vspace{-6mm}
    \caption{\textbf{Our proposed PTR-Net baseline} consists of a temporal encoder, spatial-temporal graph convolution, and attention pooling, then uses a score regressor to \textbf{predict the human-likeness score.}}
    \label{fig:baseline}
    \vspace{-4mm}
\end{figure}

We propose Pose-Temporal Regression Network (PTR-Net) for the motion human-likeness assessment tasks. It consists of three main components, and the pipeline of PTR-Net is shown in Fig.~\ref{fig:baseline}.  

\textbf{Temporal Encoder.} A two-layer bidirectional LSTM captures long-range temporal dependencies and outputs temporally encoded features \(\mathbf{H}_t \in \mathbb{R}^{2h}\). 

\textbf{Spatial–Temporal Graph Convolution (ST-GCN).} The encoded sequence is reshaped into a graph representation of the human body and processed through a stack of ST-GCN blocks. Each block alternates spatial graph convolution and temporal convolution to extract coordination patterns across joints and frames. Unlike conventional skeleton-based GCNs, PTR-Net adopts a parameter-free adjacency design, allowing more adaptive feature aggregation. 

\textbf{Attention Pooling and Regression Head.} A temporal attention module highlights salient motion segments, followed by a lightweight MLP regressor that outputs a scalar human-likeness score. 
Formally, PTR-Net learns an end-to-end mapping:
\begin{equation}
    s = f_\theta(\mathbf{X}),
\end{equation}
where \(f_\theta\) denotes the network parameters and \(s \in [0,5]\) is the predicted score (see more PTR-Net details in Supplementary File).

\textbf{Training Objective.}
PTR-Net is optimized with an L2 regression loss and a regularization term:
\begin{equation}
    \mathcal{L} = \|\hat{s} - s^*\|_2^2 + \lambda \, \mathcal{L}_{\text{reg}},
\end{equation}
\noindent where \(s^*\) denotes the human-likeness score, \(\hat{s}\) denotes the predicted score and \(\mathcal{L}_{\text{reg}}\) penalizes excessive temporal fluctuations in the predicted scores to encourage smoothness and stability.

\subsection{Evaluation Metrics}
\label{sec:metrics}
% To quantitatively assess model alignment with human perception, we define three evaluation metrics for all benchmarked models, including PTR-Net and VLM-based evaluators.

% \textbf{Mean Absolute Error (MAE).}
% MAE measures the absolute difference between the predicted scores and the human preference annotations, defined as:
% \begin{equation}
% \text{MAE} = \frac{1}{N}\sum_{i=1}^{N}|\hat{s}_i - s_i^*|,
% \end{equation}
% where \(\hat{s}_i\) and \(s_i^*\) denote the predicted and human scores for the \(i\)-th motion sequence, respectively.
% A lower MAE indicates closer numerical alignment to human judgments.

% \textbf{Root Mean Squared Error (RMSE).}
% RMSE provides a more penalized metric for large deviations, given by:
% \begin{equation}
% \text{RMSE} = \sqrt{\frac{1}{N}\sum_{i=1}^{N}(\hat{s}_i - s_i^*)^2}.
% \end{equation}
% Compared with MAE, RMSE emphasizes stability and robustness by amplifying large prediction errors.

% \textbf{Spearman’s Rank Correlation (\(\rho\)).}
% Because human-likeness scores inherently represent ordinal preferences, we also compute Spearman’s rank correlation to evaluate monotonic consistency between model predictions and human annotations:
% \begin{equation}
% \rho = 1 - \frac{6\sum_i d_i^2}{N(N^2-1)},
% \end{equation}
% where \(d_i\) is the rank difference between predicted and human scores.
% A higher \(\rho\) indicates that the model better preserves human-like relative ordering across motion samples.

To quantitatively assess how well model predictions align with human judgement, we employ three complementary metrics: Mean Absolute Error (MAE), Root Mean Squared Error (RMSE), and Spearman’s Rank Correlation (\(\rho\)).
% MAE and RMSE respectively measure absolute and squared deviations between model and human scores, reflecting numerical alignment and stability.Spearman’s \(\rho\) evaluate monotonic consistency between model predictions and human preference. 
Detailed formulations of these metrics and uses are provided in the Supplementary File.
\vspace{-2mm}

%% file: sec/experiment.tex
\section{Experiments}
% \vspace{-2mm}
In this section, we evaluate the performance of the proposed baseline and others on the motion human-likeness task. 
\label{sec:exp}
% \vspace{-2mm}
\subsection{Experimental Setup}
%Implementation Details:、Evaluation Metrics
We evaluate both our proposed pose-based baseline model and the Gemini-2.5 Pro~\cite{comanici2025gemini} and Qwen3-vl-plus~\cite{bai2023qwen} on the Motion Turing Test benchmark; it serves as a representative SOTA VLMs capable of processing video sequences. 
% We design four prompting strategies to progressively test the model’s ability to reason about human-likeness in motion sequences. 
Each SMPL-X based motion clip is fed into the model for evaluation. The baseline PTR-Net is trained on the HHMotion dataset (see training details in Supplementary File). 

% The model is trained using human annotations as supervision, optimized with Adam~\cite{adam} for 200 epochs. We use a training/validation/test split of 70\%/15\%/15\%, ensuring balanced distributions across motion categories and source domains. To quantitatively assess the alignment between model predictions and human judgments, we employ three \textbf{evaluation metrics}: MAE, RMSE and Spearman’s Rank Correlation ($\rho$). 
% \vspace{-2mm}
\subsection{VLM Human-likeness Assessment Baselines}
\label{sec:llm}
% \vspace{-1mm}
Beyond our trainable baseline, we further investigate whether VLMs can serve as evaluators for motion human-likeness. We employ Gemini 2.5 Pro~\cite{comanici2025gemini} and Qwen3-vl-plus~\cite{bai2023qwen} as VLM evaluators. The model receives rendered SMPL-X motion videos as input and rates them on a 0–5 human-likeness scale.
% , where 0 denotes completely robotic motion and 5 denotes fully human-like motion. 
% To explore the model’s reasoning capability and sensitivity to different prompting strategies, we design four evaluation settings: Zero-Shot, One-Shot, Few-Shot~\cite{brown2020language}, and \textbf{Posture-Aware} Chain-of-Thought (\textbf{PA}CoT).
To explore the model’s reasoning capability, we design four different prompting strategies: Direct Evaluation \textbf{(DE)}, Context-Guided Evaluation \textbf{(CGE)}, Prototype-Driven Evaluation \textbf{(PDE)}~\cite{brown2020language}, Direct Evaluation Chain-of-Thought (\textbf{DE}-CoT), and \textbf{Posture-Aware} Chain-of-Thought (\textbf{PA}-CoT).

%In the zero-shot setting, the model directly evaluates the target motion based solely on the instruction, without any demonstration. The One-Shot setup introduces a single example with a reference score and a descriptive rationale to guide the model’s understanding of evaluation style. The few-shot setup provides six labeled examples covering all score levels (0–5), each paired with a human-written description. This design enables the model to observe a range of motion qualities and learn the subtle boundaries between human-like and robotic behaviors, aiming to better approximate human scoring tendencies. The setting details are provided in Appendix 1.
In the DE method, the VLM directly evaluates the target motion solely based on the instruction, without any demonstration. The CGE setup introduces a single example with a reference score and a description to guide the VLM’s understanding of evaluation style. The PDE setup provides six labeled examples covering all score levels (0–5) to calibrate the human-likeness score, each paired with a human-written description. The DE-CoT method introduces a CoT approach~\cite{wei2022chain}, enabling the model to reason step-by-step to reason the result. The PA-CoT method we proposed adopts a structured reasoning process that mirrors human evaluators’ thought patterns. The model is first instructed to identify the motion type (upper limb, lower limb, or full-body) and then to analyze the motion along three interpretable dimensions: \textbf{posture fluency}, \textbf{movement coordination}, and \textbf{core stability}. Based on these analyses, the model synthesizes a final score and generates a qualitative description.

%在零样本设置中，模型仅依据指令直接评估目标动作，无需任何示范。单样本设置引入了一个示例，并附有参考分数（例如 3 分）以及一段描述性的解释，以指导模型理解评估方式。少样本设置提供了六个已标注的示例，涵盖了所有评分等级（0 至 5），每个示例都配有人工撰写的描述。这种设计使模型能够观察到各种动作质量，并学习人类行为与机器人行为之间细微的界限，旨在更接近人类评分倾向。

 %This structured reasoning encourages explainable and human-aligned evaluation, enabling the model to capture subtle differences in motion smoothness and balance that are often overlooked in direct scoring.

%Comparative results of different methods are reported in Sec.\ref{sec:exp}, showing that the PACoT few-shot demonstration achieves the highest consistency with human ratings, outperforming zero-shot and standard few-shot approaches. We evaluate all models using the metrics described in Sec.~\ref{sec:metrics}.
%Notably, Qwen3-VL-Plus exhibits near-identical outputs across the DE, CGE, PDE, and PACoT settings. %, reflecting its weaker temporal and contextual reasoning ability compared with Gemini.
%Notably, due to model limitations, Qwen3-VL-Plus exhibits near-identical outputs across the DE, CGE, PDE, and PACoT settings, reflecting its weaker temporal and contextual reasoning ability compared with Gemini.

\subsection{Quantitative Results}

\begin{table}[t]
\centering
\caption{\textbf{Quantitative results for different models} on the Motion Turing Test benchmark. “*” indicates VLM-based evaluation without task-specific training. Qwen3-VL-Plus (shot)* denotes that its results remain identical in DE/CGE/PED/DE-CoT/PA-CoT settings. “--” indicates invalid $\rho$ due to constant outputs.}
\vspace{-2mm}
\resizebox{\linewidth}{!}{
\begin{tabular}{lccc}
\toprule
\textbf{Model} & \textbf{MAE} $\downarrow$ & \textbf{RMSE} $\downarrow$ & \textbf{Spearman's $\rho$} $\uparrow$ \\
\midrule

% \multicolumn{4}{c}{\textit{General-Purpose VLMs (No Task-Specific Training)}} \\
% \midrule
Gemini 2.5 Pro (DE)* & 1.3105 & 1.5873 & 0.1609 \\
Gemini 2.5 Pro (CGE)* & 1.3314 & 1.5986 & 0.1658 \\
Gemini 2.5 Pro (PDE)* & 1.2616 & 1.5397 & 0.2188 \\
Gemini 2.5 Pro (DE-CoT)* & 1.6040 & 1.9146 & 0.1353 \\
Gemini 2.5 Pro (PA-CoT)* & 1.2682 & 1.5214 & 0.2303 \\
Qwen3-VL-Plus (shot)* & 1.7714 & 2.1018 & -- \\

\midrule
% \multicolumn{4}{c}{\textit{Motion-Based and Lightweight Baselines}} \\
% \midrule
MotionBERT (Frozen Backbone) & 0.6846 & 0.9025 & 0.5315 \\
MotionBERT (Fine-tuned) & 0.6252 & 0.8465 & 0.6142 \\
Transformer (Lightweight) & 0.6387 & 0.8259 & 0.5728 \\

\midrule
\textbf{PTR-Net (Ours)} & \textbf{0.5813} & \textbf{0.7926} & \textbf{0.6841} \\

\bottomrule
\end{tabular}
}
\label{tab:quantitative_results}
\vspace{-2mm}
\end{table}

The quantitative comparison is shown in Tab.~\ref{tab:quantitative_results}. Despite the impressive reasoning capabilities of recent LLMs, assessing human-likeness in motion remains highly challenging. Even with structured strategies such as PA-CoT, Gemini 2.5 Pro exhibits large deviations from human judgments.
Qwen3-VL-Plus shows even lower sensitivity to motion details, yielding nearly identical predictions across different setups and a higher error. 
%To address this gap, we propose a simple yet effective baseline model that can directly learn to regress human-likeness scores from pose sequences.
To provide stronger motion-centric baselines, we further compare PTR-Net with MotionBERT~\cite{MotionBert2023} under both frozen and fine-tuned settings, as well as a Transformer-based~\cite{vaswani2017attention} lightweight baseline.

% Overall, PTR-Net consistently outperforms all baselines across metrics, demonstrating that it serves as a strong yet efficient model for the human-likeness evaluation task under fair task-specific training. However, the 0.80 RMSE indicates that there is still room for improvement for the motion human-likeness task. 

Overall, PTR-Net consistently outperforms all baselines across metrics, achieving lower MAE and RMSE and higher $\rho$. However, the 0.80 RMSE indicates that there is still room for improvement in this task. 

\begin{figure}[t]
% \vspace{-2mm}
    \centering
    \includegraphics[width=1\linewidth, height=0.9\linewidth]{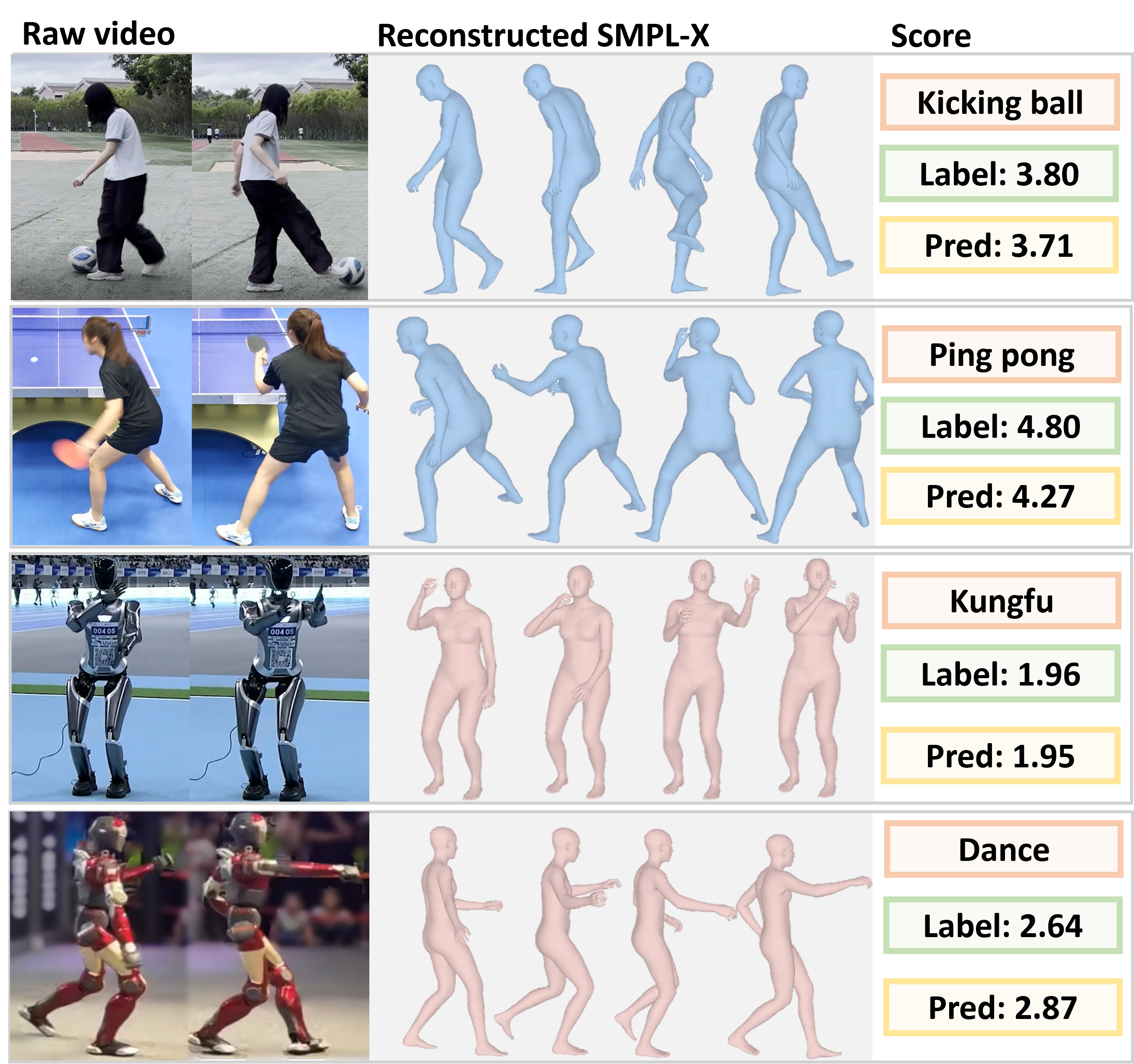}
    \vspace{-6mm}
    \caption{Four representative motion examples with \textbf{PTR-Net predictions compared to human-annotated scores.}}
    \label{fig:qualitative}
    \vspace{-6mm}
\end{figure}

% \vspace{-2mm}
\subsection{Qualitative Evaluation}
\label{sec:qualitative}

To further illustrate the interpretability and reliability of PTR-Net’s evaluation behavior, we visualize four representative examples across both human and humanoid motion types in Fig.~\ref{fig:qualitative}, PTR-Net’s predictions exhibit strong alignment with human-likeness scores across diverse categories.

For human actions, PTR-Net’s predictions closely align with human evaluations. In kicking ball (label 3.8 / pred 3.71) and playing ping pong (label 4.8 / pred 4.27), the model correctly captures smooth coordination and natural limb transitions. For humanoid actions, it accurately identifies mechanical rigidity. PTR-Net reflects human judgments by assigning lower scores due to visible stiffness and rhythmic discontinuities.
Overall, these examples show that the proposed PTR-Net not only matches human ratings numerically but also captures interpretable motion cues such as fluency, coordination, and balance.

% \vspace{-2mm}
\subsection{Ablation Study}
\label{sec:ablation}

\begin{table}[t]
% \vspace{-4mm}
\centering
\caption{\textbf{Ablation study of PTR-Net components} on the Motion Turing Test benchmark.}
\vspace{-1mm}
\resizebox{\linewidth}{!}{
\begin{tabular}{lccc}
\toprule
\textbf{Model Variant} & \textbf{MAE} $\downarrow$ & \textbf{RMSE} $\downarrow$ & \textbf{Spearman's $\rho$} $\uparrow$ \\
\midrule
w/o Temporal Encoder & 0.7631 & 0.9691 & 0.3610 \\
w/o Attention Pooling & 0.6185 & 0.8203 & 0.6255 \\
w/o $\mathcal{L}_{reg}$ & 0.5983 & 0.7958 & 0.6215 \\
\midrule
\textbf{PTR-Net(Full Model)} & \textbf{0.5813} & \textbf{0.7926} & \textbf{0.6841} \\
\bottomrule
\end{tabular}
}
\label{tab:ablation}
\vspace{-4mm}
\end{table}

To better understand the contribution of each component in PTR-Net, we conduct an ablation study by selectively removing or modifying key modules. As shown in Tab.~\ref{tab:ablation}, removing the temporal encoder results in an increase in MAE and RMSE errors. Disabling attention pooling also reduces correlation with human ratings, indicating its role in capturing global temporal context. The regularization term $\mathcal{L}_{reg}$ further stabilizes training and improves score consistency. The full model achieves the optimal overall balance between error reduction and ranking consistency, verifying the rationality of the collaborative design of each component.

\subsection{Discussion on Humans Imitating Humanoids}
\label{sec:imitate}
\begin{figure}[t]
\vspace{-2mm}
    \centering
    \includegraphics[width=1.0\linewidth, height=0.4\textwidth]{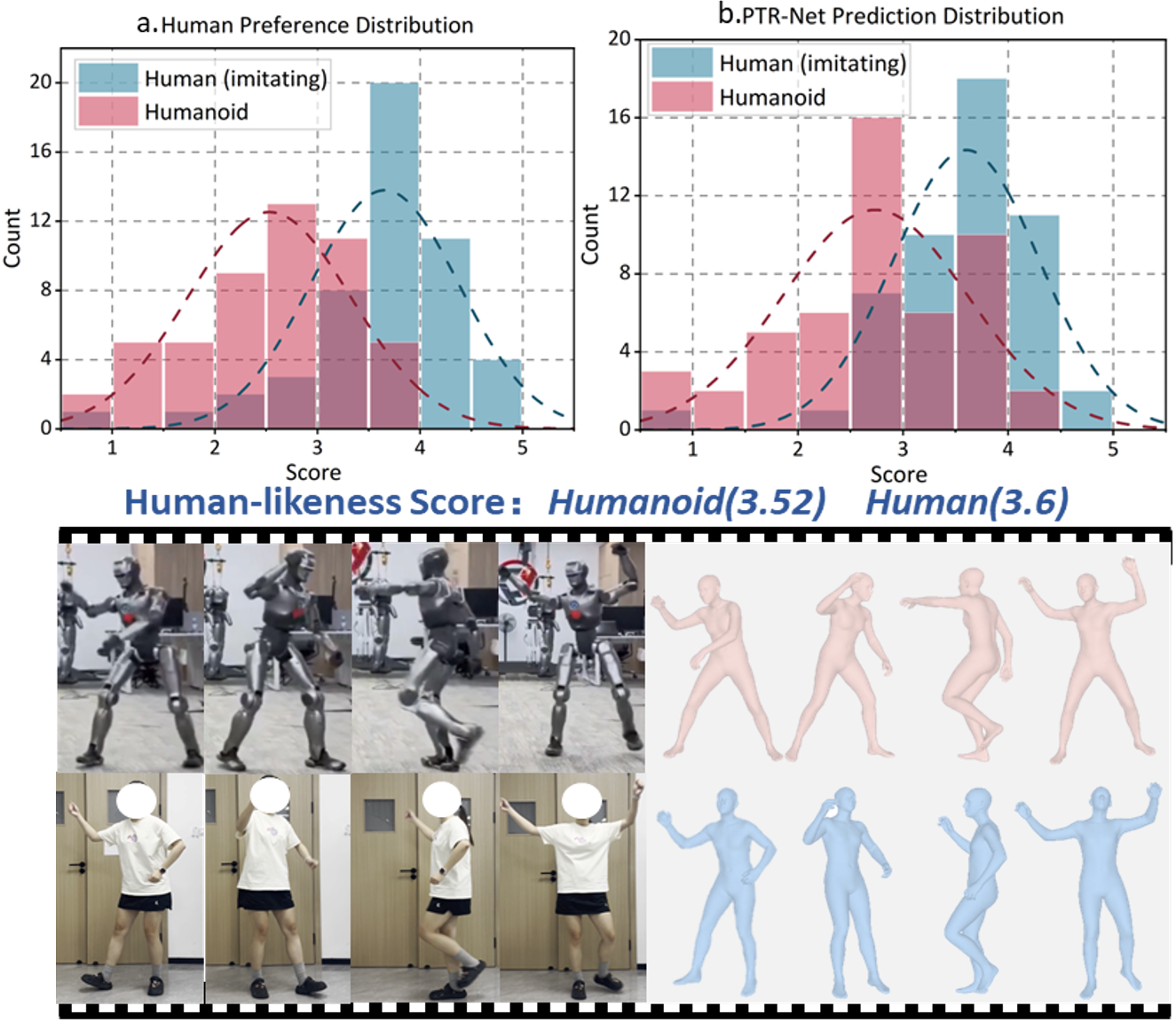}
    \vspace{-6mm}
    \caption{(a–b) Score distributions of \textbf{human annotators and PTR-Net predictions} for both \textbf{human (imitating humanoid) and humanoid motions}. (Bottom) Representative dance imitation pairs illustrating near-indistinguishable motion patterns.}
    \label{fig:imitate}
    % \vspace{-2mm}
\end{figure}
We analyze the \textit{human-imitating-humanoid} subset to examine evaluation boundaries of motion human-likeness. As shown in Fig.~\ref{fig:imitate}(a–b), PTR-Net closely follows human annotators' distributions, suggesting its ability to capture subtle temporal and coordination cues. Human ratings show partial overlap between imitating humans and humanoids, indicating that intentional imitation can blur the evaluation boundary of human-likeness.

The bottom dance imitation pairs further support this observation, that both human and humanoid motions receive nearly identical scores. These findings highlight a \textbf{non-trivial} ambiguity in human-likeness evaluation: when humans intentionally mimic humanoids' mechanical rigidity, the kinematic cues become insufficient for reliable discrimination. This suggests that true human-likeness involves not only smoothness and coordination but also the intentionality and adaptability underlying human motion dimensions, which remain challenging for both the generation and model assessment of humanoids.

\subsection{Out-of-Distribution Evaluation}
\begin{figure}[t]
    \centering
    \includegraphics[width=1\linewidth]{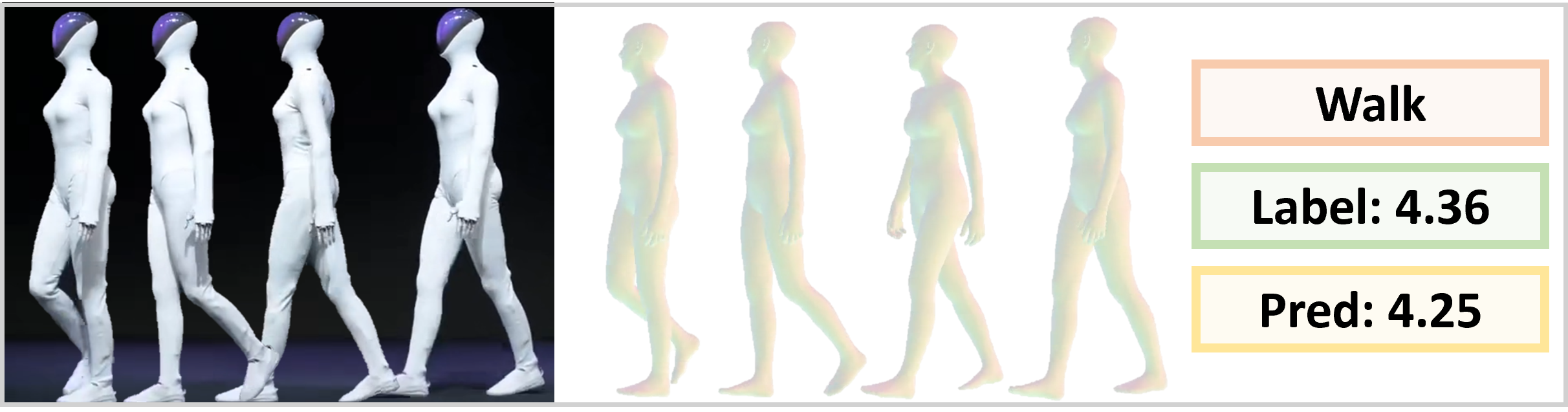}
    \vspace{-6mm}
    \caption{\textbf{Out-of-distribution evaluation} on the \textbf{XPeng IRON~\cite{xiaopeng}} humanoid robot.}
    \label{fig:xiaopeng}
    \vspace{-6mm}
\end{figure}
To examine the generalization ability of our benchmark beyond the distribution of the collected data, we further tested it on motions from unseen humanoid robots, including the recently released (Nov. 2025) XPeng IRON~\cite{xiaopeng} as shown in Fig.~\ref{fig:xiaopeng}. 
PTR-Net predicts an outstanding performance of the XPeng IRON human-likeness score of 4.25, closely matching the human annotation average of 4.36. 
% from 25 evaluators (the IAC check removed five inconsistent annotations). 
This strong alignment not only demonstrates the robustness of PTR-Net in handling unseen humanoid models but also validates the consistency and reliability of motion human-likeness scoring. More OOD examples will be added as they become open-sourced (see more examples in Supplementary File).
% and several other contemporary models. 
% Comprehensive results and visual examples are provided in the Appendix.

% \section{Analysis and Discussions}
% \label{sec:discuss}